# A Generative Bayesian Model for Aggregating Experts' Probabilities

**Joseph M. Kahn**
Department of Management Science & Engineering
Stanford University
Stanford, CA 94305
jkahn@stanford.edu

## Abstract

In order to improve forecasts, a decision-maker often combines probabilities given by various sources, such as human experts and machine learning classifiers. When few training data are available, aggregation can be improved by incorporating prior knowledge about the event being forecasted and about salient properties of the experts. To this end, we develop a generative Bayesian aggregation model for probabilistic classification. The model includes an event-specific prior, measures of individual experts' bias, calibration, accuracy, and a measure of dependence betweeen experts. Rather than require absolute measures, we show that aggregation may be expressed in terms of *relative* accuracy between experts. The model results in a weighted logarithmic opinion pool (LogOps) that satisfies consistency criteria such as the external Bayesian property. We derive analytic solutions for independent and for exchangeable experts. Empirical tests demonstrate the model's use, comparing its accuracy with other aggregation methods.

## 1 Introduction

It is natural to consult one or more experts for better information when making a decision under uncertainty. Forecasts may come from human experts and from computer models — for instance, as the result of learning probabilistic classifiers. To improve accuracy, many methods have been proposed to combine such forecasts. This problem has been studied under the labels "expert aggregation" and "opinion pooling" [Genest & Zidek, 1986; Cooke, 1991; Clemen & Winkler, 1999], "ensemble learning," and "mixture of experts" [Dietterich, 2003; Jacobs et al., 1991].

A difficulty of complex models is that they require large data sets for training, not typically available with human experts. Indeed, prevailing wisdom is that simple averages usually outperform more complex models [Wallsten et al., 1997]. In lieu of massive training data, a decision maker (DM) may be called upon to assess model parameters. If we expect to improve upon simple averaging, a practical model should strike a balance: it should capture key, easily interpretable characteristics of experts while using a minimal number of parameters; yet it should still offer enough flexibility to take advantage of large training sets when available.

In this paper we examine probability forecasts to classify a *categorical* target $Y \in \mathcal{Y} = \{1, 2, \cdots, K\}$. Given a set of $N$ stated probabilities $\mathbf{q} \equiv (q^{(1)}, q^{(2)}, \cdots q^{(N)})$ on $\mathcal{Y}$, the goal is to assign an aggregate forecast $p^* = g(\mathbf{q})$. *Discriminative* approaches *directly* learn $g(\mathbf{q})$. A functional form is assigned to $g$, and parameters (usually weights) are estimated. Two popular functional forms are linear opinion pooling (LinOps) and logarithmic opinion pooling (LogOps):

$$p^{LinOps}(y) \equiv \sum_{i=1}^{N} w_i q^{(i)}(y); \qquad (1)$$

$$L_{y:z}^{LogOps} \equiv w_o + \sum_{i=1}^{N} w_i L_{y:z}^{(i)}. \qquad (2)$$

LogOps is expressed in terms of the log odds between two classes $y, z \in \mathcal{Y}$:

$$L_{y:z}^{(i)} \equiv \log \frac{q^{(i)}(y)}{q^{(i)}(z)}. \qquad (3)$$

LogOps weights are unconstrained, and their direct estimation is equivalent to logistic regression. LinOps weights are constrained $0 \le w_i \le 1$, $\sum_i w_i = 1$, forming a convex combination of probabilities.

In contrast to the discriminative approach, *generative* methods *jointly* learn a distribution for $\mathbf{q}$ and $Y$, and then infer $Y$ given $\mathbf{q}$. Under the generative Bayesian approach [Morris, 1971], a decision-maker (DM) first assesses her prior probability $p^o$ on $\mathcal{Y}$. She also assesses the conditional likelihood function, $f_{\mathbf{Q}|Y}(\mathbf{q}|y)$. Then she infers $p^*$ using Bayes' rule:

$$p^*(y) \equiv f_{Y|\mathbf{Q}}(y|\mathbf{q}) = \frac{f_{\mathbf{Q}|Y}(\mathbf{q}|y)p^o(y)}{\sum_{z \in \mathcal{Y}} f_{\mathbf{Q}|Y}(\mathbf{q}|z)p^o(z)}. \quad (4)$$

Use of a functional form and simplifying assumptions ease assessment of the likelihood function. One oft-used simplification is the independent opinion pool (IndOps), which assumes experts to be conditionally independent given $Y$. This simplification is equivalent to a product-of-experts [Hinton, 1999] or "naïve Bayes" classifier [Langley, 1994]. IndOps results in a LogOps formula, but with each of the experts' weights set to unity.

Beyond IndOps, French [1981] and Lindley [1982] present generative Bayesian models where the likelihood function for experts' log-odds follows a Gaussian distribution. In the following sections, we specialize these models to yield a LogOps aggregation rule, akin to Fisher's [1936] classical linear discriminant analysis. We express the entire model with parameters that *directly* quantify experts' bias, calibration, accuracy, and dependence. Further, we derive closed form solutions for the special cases of conditionally-independent experts, and for exchangeable experts. Finally, we test the model empirically, comparing its accuracy to commonly-used alternatives.

## 2 Gaussian model

### 2.1 Model formulation

A decision maker (DM) has event prior $p^o$ and corresponding log-odds prior $L^o$ on categorical event $Y \in \mathcal{Y}$. Due to space limitations, we limit this derivation to the binomial case $\mathcal{Y} = \{0, 1\}$. (Kahn [2004] extends $\mathcal{Y}$ to multinomial events.) The set of $N$ experts' forecasts are given as the vector of log-odds $\mathbf{L} \equiv \left[ L_{1:0}^{(1)}, \cdots, L_{1:0}^{(N)} \right]^T$. We use the simplifying assumption of Gaussian distributions with constant conditional (on class $Y$) covariance matrices:

$$\mathbf{L}_{1:0}|Y = 1 \sim \mathcal{N}(\boldsymbol{\mu}_1, \boldsymbol{\Sigma}), \; \mathbf{L}_{1:0}|0 \sim \mathcal{N}(\boldsymbol{\mu}_0, \boldsymbol{\Sigma}). \quad (5)$$

We define *bias* as the expected difference

$$\mathbf{b}_{1:0} \equiv \frac{1}{2}(\boldsymbol{\mu}_1 + \boldsymbol{\mu}_0) - L_{1:0}^o \mathbf{1}, \quad (6)$$

and conveniently work with *debiased contrasts*

$$\mathbf{C}_{1:0} \equiv \mathbf{L}_{1:0} - L_{1:0}^o \mathbf{1} - \mathbf{b}_{1:0}. \quad (7)$$

Note that from these definitions it immediately follows that $\mathbf{L}_{0:1} = -\mathbf{L}_{1:0}$, $\mathbf{b}_{0:1} = -\mathbf{b}_{1:0}$, and $\mathbf{C}_{0:1} = -\mathbf{C}_{1:0}$, so that with contrast mean

$$\boldsymbol{\mu} \equiv \frac{1}{2}(\boldsymbol{\mu}_1 - \boldsymbol{\mu}_0), \quad (8)$$

we have

$$\mathbf{C}_{1:0}|1 \sim \mathcal{N}(\boldsymbol{\mu}, \boldsymbol{\Sigma}), \quad \mathbf{C}_{0:1}|0 \sim \mathcal{N}(\boldsymbol{\mu}, \boldsymbol{\Sigma}). \quad (9)$$

In our model, $p^o$ represents common background information between the DM and all experts. If we wanted to represent a DM having relevant *private* knowledge, we would still specify $p^o$ as a common reference prior; but we would also include the DM's (non-reference) prior as an additional expert.

### 2.2 Model solution

Defining the weight vector

$$\boldsymbol{w} \equiv 2\boldsymbol{\Sigma}^{-1}\boldsymbol{\mu}, \quad (10)$$

we calculate the aggregate contrast via Bayes' rule in log-odds form:

$$C_{1:0}^* \equiv L_{1:0}^* - L_{1:0}^o \equiv \log\left(\frac{f_{Y|\mathbf{C}}(1|\mathbf{C}_{1:0})}{f_{Y|\mathbf{C}}(0|\mathbf{C}_{1:0})}\right) - L_{1:0}^o$$

$$= \log\left(\frac{f_{\mathbf{C}|Y}(\mathbf{C}_{1:0}|1)}{f_{\mathbf{C}|Y}(\mathbf{C}_{1:0}|0)}\right) = -\frac{1}{2}(\mathbf{C}_{1:0}-\boldsymbol{\mu})^T\boldsymbol{\Sigma}^{-1}(\mathbf{C}_{1:0}-\boldsymbol{\mu})$$

$$-\frac{1}{2}(-\mathbf{C}_{1:0}-\boldsymbol{\mu})^T\boldsymbol{\Sigma}^{-1}(-\mathbf{C}_{1:0}-\boldsymbol{\mu}) = \boldsymbol{w}^T\mathbf{C}_{1:0}. \quad (11)$$

Defining $L^{(0)} \equiv L^o$, $w^{(0)} \equiv 1 - \sum_{i=1}^N w^{(i)}$ and $b^{(0)} \equiv 0$, we can rewrite equation (11) as a (de-biased) LogOps:

$$L_{1:0}^* = \sum_{i=0}^N w^{(i)}(L_{1:0}^{(i)} - b_{1:0}^{(i)}). \quad (12)$$

A type of unanimity holds in that $L^*$ matches the debiased forecasts when all (including $L^o$) are in agreement. Since the weights on all forecasts (including $L^o$) in this LogOps total unity, this aggregation rule satisfies the "externally Bayesian" criterion [Genest et al., 1986].

## 3 Feature abstraction

An advantage of the model is that its parameters are readily interpreted as meaningful features of individual and pairwise experts. The prior log odds $L^o$ allow us to represent a baseline level of prediction difficulty. $b^{(i)}$ expresses unwarranted individual bias away from baseline probabilities; when $L^{(i)} = L^o + b^{(i)}$, the aggregate remains unchanged at $L^o$. Dependence between experts $i$ and $j$ is represented by the usual Pearson's correlation coefficient $\rho_{(i,j)} \equiv \frac{\boldsymbol{\Sigma}_{i,j}}{\sigma_{(i)}\sigma_{(j)}}$, where $\sigma_{(i)}^2 \equiv \boldsymbol{\Sigma}_{i,i}$ is expert $i$'s individual forecast variance.

### 3.1 Calibration

Apart from any bias towards specific classes, *calibration* is a term used to describe how accurately an expert can encode his knowledge as a probability forecast. It can be thought of as a measure of an expert's over-confidence (regardless of class), stating forecasts that are too extreme for his knowledge. (We could think of such a situation arising for a machine learning classifier due to overfitting).

If we use *only* expert $i$, then the weight in equation (10) is interpretable as a *calibration coefficient* [Lindley, 1982]:

$$\kappa^{(i)} \equiv 2\frac{\mu^{(i)}}{\sigma^2_{(i)}}, \tag{13}$$

and then $C^{*(i)} = \kappa^{(i)}C^{(i)}$, or

$$L^{*(i)} = \kappa^{(i)}(L^{(i)} - b^{(i)}) + (1 - \kappa^{(i)})L^o. \tag{14}$$

If $\kappa^{(i)} = 1$ and we are given *only* expert $i$'s forecast we would adopt it "as is." An overconfident expert has $\kappa^{(i)} < 1$, leading us to shade $C^{*(i)}$ to a less extreme contrast. $\kappa^{(i)} > 1$ represents an underconfident expert, leading us to increase the expert's contrast. If $\kappa^{(i)} = 0$, then $C^{*(i)} = 0$, retaining the reference prediction $L^o$. A *calibration curve* shows the relation between a single expert's stated probability and the resulting update. In figure 1 below, we construct calibration curves for use of a single, unbiased ($b = 0$) expert, on a binomial event (with $p^o = 0.3$) at various levels of $\kappa$.

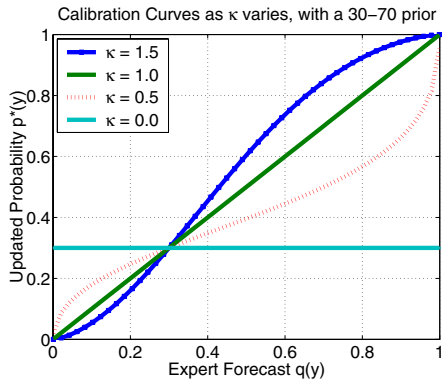

Figure 1: Calibration curves for unbiased experts.

### 3.2 Accuracy

While the vector $\boldsymbol{\mu}$ is a measure of contrast in the experts' *stated* forecasts, we define $\mu^*$ as their combined accuracy when optimally aggregated:

$$\mu^* \equiv \mathbf{E}\{C^*_{Y:Y'}\} = \sum_{y \in \mathcal{Y}} p^o(y)\mathbf{E}\{C^*_{y:y'}|y\}$$
$$= \sum_{y \in \mathcal{Y}} p^o(y)\mathbf{E}\{\boldsymbol{w}^T\mathbf{C}_{y:y'}|y\} = \boldsymbol{w}^T\boldsymbol{\mu}. \tag{15}$$

If we use but a single expert $i$, accuracy is the product of individual calibration and contrast: $\mu^{*(i)} = \kappa^{(i)}\mu^{(i)}$. We can thus write the contrast parameter as $\mu^{(i)} = \frac{\mu^{*(i)}}{\kappa^{(i)}}$, a function of individual accuracy and calibration.

### 3.3 Relative accuracy of experts

Equation (10) expresses weights $\boldsymbol{w}$ as a function of $\boldsymbol{\mu}$, the experts' contrast means. A downside of this representation is that $\boldsymbol{\mu}$ is specific to the difficulty of forecasting the particular event: for some events, all of the experts may have much less information than for other events. We should have greater confidence that a *relative* measure of accuracy between experts could be successfuly assessed. To this end, we define the *relative accuracy* as a ratio:

$$m^{(i)} \equiv \frac{\mu^{*(i)}}{\mu^{*(1)}}. \tag{16}$$

We now prove that our models' weights may be expressed in terms of relative accuracy, without resorting to an absolute measure of contrast.

**Theorem 3.1 (Relative weights)** *The aggregation weights in equation (10) may be written as a function of experts' calibration $\boldsymbol{\kappa}$, correlation matrix $\mathbf{R}$, and relative accuracy $\mathbf{m}$, but not a direct function of absolute contrast $\boldsymbol{\mu}$.*

**Proof** Using $\kappa^{(i)} \equiv \frac{2\mu^{(i)}}{\sigma^2_{(i)}}$ and $\mu^{*(i)} = \kappa^{(i)}\mu^{(i)}$ yields

$$\frac{1}{\sigma_{(i)}} = \sqrt{\frac{\kappa^{(i)}}{2\mu^{(i)}}} = \frac{\kappa^{(i)}}{\sqrt{2\mu^{*(i)}}} = \frac{\kappa^{(i)}}{\sqrt{2m^{(i)}\mu^{*(1)}}}; \tag{17}$$

$$\frac{\mu^{(i)}}{\sigma_{(i)}} = \frac{\kappa^{(i)}\mu^{(i)}}{\sqrt{2\mu^{*(i)}}} = \sqrt{\frac{\mu^{*(i)}}{2}} = \sqrt{\frac{m^{(i)}\mu^{*(1)}}{2}}. \tag{18}$$

Then with $\mathbf{D} \equiv diag\left[\sigma^2_{(1)}, \cdots, \sigma^2_{(N)}\right]$, we have

$$\boldsymbol{w}^T = 2\boldsymbol{\mu}^T\boldsymbol{\Sigma}^{-1} = 2\boldsymbol{\mu}^T\mathbf{D}^{-\frac{1}{2}}\mathbf{R}^{-1}\mathbf{D}^{-\frac{1}{2}}$$
$$= (\mathbf{m}^T)^{-\frac{1}{2}}\mathbf{R}^{-1}diag\left[\frac{\kappa^{(1)}}{\sqrt{m^{(1)}}}, \cdots, \frac{\kappa^{(N)}}{\sqrt{m^{(N)}}}\right]. \quad \blacksquare \tag{19}$$

## 4 Special cases

### 4.1 Recalibration of a single expert

A common strategy for classification with many covariates is a hierarchical mixture of experts. "Weak learners" are first trained on a single covariate, and then their predictions aggregated. If a covariate is categorical, a probability can be estimated from count data. If a covariate is continuous, we may regard it as a distorted log-odds statement that we wish to recalibrate and de-bias: assess or estimate the prior $L^o$, individual conditional means $\mu_0^{(i)}, \mu_1^{(i)}$, and individual conditional variance $\sigma_{(i)}^2$ for each learner; then apply equation (14) to de-bias and recalibrate. For each learner $(i)$, the updated contrast mean given $Y = y$ is

$$\mu_y^{*(i)} \equiv \mathbf{E}\{L_{1:0}^{*(i)}|y\} = \kappa^{(i)}(\mu_y^{(i)} - b_{1:0}^{(i)}) + (1 - \kappa^{(i)})L_{1:0}^o. \tag{20}$$

The update is unbiased, since

$$b_{1:0}^{*(i)} \equiv \frac{1}{2}(\mu_1^{*(i)} + \mu_0^{*(i)}) - L_{1:0}^o$$
$$= \frac{1}{2}\kappa^{(i)}(\mu_1^{(i)} + \mu_0^{(i)} - 2b_{1:0}^{(i)}) + (1 - \kappa^{(i)})L_{1:0}^o - L_{1:0}^o$$
$$= 0. \tag{21}$$

The update is also calibrated, since $\mu^{*(i)} = \kappa^{(i)}\mu^{(i)}$ and $\sigma_{*(i)}^2 = (\kappa^{(i)})^2\sigma_{(i)}^2$ yield $\kappa^{*(i)} \equiv \frac{2\mu^{*(i)}}{\sigma_{*(i)}^2} = 1$.

### 4.2 Aggregation of independent experts

For independent experts, $\rho_{i,j} = 0$ for all pairs of experts $i, j \neq i$. So $\mathbf{\Sigma} = diag\left[\sigma_{(1)}^2, \cdots, \sigma_{(N)}^2\right]$ and

$$\boldsymbol{w}^T = 2\mathbf{m}^T\mathbf{\Sigma}^{-1} = [\kappa^{(1)}, \cdots, \kappa^{(N)}]. \tag{22}$$

As might be expected, the case of unbiased ($b^{(i)} = 0$), calibrated ($\kappa^{(i)} = 1$) experts leads us immediately to an IndOps rule:

$$L_{1:0}^{IndOps} = (1 - N)L_{1:0}^o + \sum_{i=1}^{N} L_{1:0}^{(i)}. \tag{23}$$

(Note the $(1 - N)L_{1:0}^o$ term that is often neglected in IndOps expositions).

### 4.3 Aggregation of exchangeable experts

Another special case of the model is when there isn't enough information to distinguish between experts: $N$ experts with equal means $\mu$, variance $\sigma^2$, calibration $\kappa$, and identical pairwise correlation coefficients $\rho$. For $|\rho| < 1$, it is easily verified that $\mathbf{\Sigma}^{-1}$ has identical diagonal elements $\frac{\alpha}{\sigma^2}$, and identical off-diagonal elements $\frac{\beta}{\sigma^2}$. Here, $\alpha \equiv \left(\frac{1+(N-2)\rho}{(1-\rho)[1+(N-1)\rho]}\right)$, $\beta \equiv \left(\frac{\rho}{(1-\rho)[1+(N-1)\rho]}\right)$. Using equation (10),

$$\boldsymbol{w}^T = 2\mu\mathbf{1}^T\mathbf{\Sigma}^{-1} = \frac{2\mu}{\sigma^2}[\alpha - (N-1)\beta]\mathbf{1}^T$$
$$= \kappa\left(\frac{1}{1+(N-1)\rho}\right)\mathbf{1}^T. \tag{24}$$

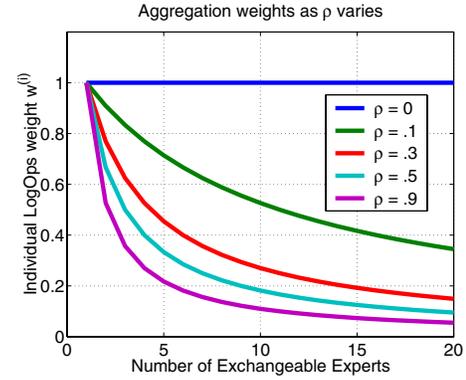

(a) Experts' aggregation weights

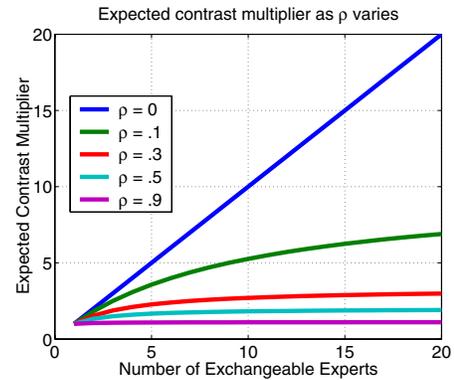

(b) Aggregate accuracy multipliers

Figure 2: Exchangeable experts' weights and accuracy multipliers.

Note that for independent experts ($\rho = 0$), weights for *each* expert in the aggregation are $w^{(i)} = \kappa$, the same weight as when $i$ is used alone. For redundant experts ($\rho = 1$), the weights are $w^{(i)} = \frac{\kappa}{N}$, and the *sum* of all $N$ weights remains fixed at $\kappa$. Combining equations (15) and (24), the aggregate accuracy using experts $(1, \cdots, N)$ is a multiple of the single-expert accuracy:

$$\mu^{*(1,\cdots,N)} = \boldsymbol{w}^T\mathbf{m} = \left(\frac{N}{1+(N-1)\rho}\right)\mu^{*(1)}. \tag{25}$$

Individual weights and the aggregate *accuracy multiplier* are plotted in figure 2 for calibrated, exchangeable experts. The plots show the effects of increasing both $N$ and $\rho$.

For independent ($\rho = 0$) experts, the aggregate contrast in equation (25) is the multiple $N$ of one individual's contrast. For large $N$, the contrast multiplier approaches $\frac{1}{\rho}$, which places a clear bound on the informativeness of a group of dependent experts. In extreme cases, a large number of replicated experts ($\rho = 1$) yields no more accuracy than a single expert, while aggregating many independent experts yields near perfect information.

Assuming near-calibrated ($\kappa \approx 1$) experts, the aggregate for high $\rho$ is typically quite close numerically to a LinOps combination. For example, a binomial event with $p^o(y) = 50\%$, and two exchangeable experts with $\rho = 0.95$, $\kappa = 0.9$ yields $w^{(i)} = 0.51$. If $q^{(1)}(y) = 70\%$ and $q^{(2)}(y) = 80\%$, their aggregate is a compromise $p^*(y) = 76\%$. If instead $\rho = 0$ then the forecasts reinforce each other: $w^{(i)} = 0.9$ and $p^*(y) = 88\%$.

# 5 Empirical testing

To demonstrate use of the model, we performed experiments on both simulated and actual data, comparing the accuracy scores between our Gaussian model with those of IndOps (naïve Bayes), LinOps, logistic regression (discriminative LogOps), and also a simple equal-weighted average of probabilities (AvgOps).

## 5.1 Data sets

First, we simulated three data sets matching our model's distributional assumptions of conditional Gaussian log-odds forecasts. Each set had 10 experts forecasting binomial $Y \in \{0, 1\}$, with 500 instances in each set. To focus our first experiment, we only changed $\rho$ in simulating these sets ($\rho = 0.0$ on the first set, $\rho = 0.2$ on the second, and $\rho = 0.5$ on the third). Other simulation parameters were constant over all sets and over all experts: $p(1) = 0.5$, $\mu_0^{(i)} = -1$, $\mu_1^{(i)} = +1$, $\sigma_{(i)}^2 = 1$ for all $i$. Although we don't assume it to be true when learning from training data, these simulated experts are all exchangeable, unbiased, and fully calibrated, since: $L_{1:0}^o = \log \frac{p^o(1)}{p^o(0)} = 0$, $\mathbf{b} = \frac{1}{2}(\boldsymbol{\mu}_1 + \boldsymbol{\mu}_0) + \mathbf{1}L^o = \mathbf{0}$, $\boldsymbol{\mu} = \frac{1}{2}(\boldsymbol{\mu}_1 - \boldsymbol{\mu}_0) = \mathbf{1}$, and $\kappa^{(i)} = \frac{2\mu^{(i)}}{\sigma_{(i)}^2} = 1$.

We next demonstrate aggregation on data sets from the UCI machine learning repository [Blake & Mertz, 1998]. The selected sets had a binomial target variable, numerical covariates, and complete data. We show results from the Pima Indians diabetes data (768 instances of 8 real-valued predictors and one binomial target class), and the liver disorders data from BUPA Medical Research (with 345 instances of 6 real-valued predictors and one binomial target class).

## 5.2 Randomization

For each data set, we recorded the average score on test instances as we increased the number $N$ of experts in the mix. We performed 30 iterations for each set, for each vale of $N$. In each iteration, we randomly selected (without replacement) the $N$ experts, and also half of the instances for training. The remaining half of the instances were used for testing (scoring).

## 5.3 Scoring

We evaluated the accuracy of agregate forecasts using a log scoring-rule, which is known to be, in general, the only smooth scoring rule that is simultaneously *local* and *strictly proper* [Bernardo & Smith, 1994]. That is, a score depends only on the probability assigned to the correct class, and an expert receives the highest expected score only by disclosing his true beliefs. We use a particular scaling that makes the log score easily interpretable:

$$S(q, Y; p) \equiv \frac{\log(q(Y)/p(Y))}{H(p)}, \qquad (26)$$

where $q(Y)$ is the probability to be scored, $Y$ is the outcome, $H(p) \equiv -\sum_{y \in \mathcal{Y}} p(y) \log p(y)$ is the information entropy in scoring prior $p$. This form yields a score of 0 if the expert simply matches $p$. It also normalizes the score so that — under $p$ — the expected score is unity for an expert with perfect information. To avoid comparison difficulties, we used a uniform (50-50) scoring prior for our experiments.

A difficulty of the log-score occurs with the rare statement of zero probability; in our experiments, we avoid this by enforcing a lower bound of $10^{-4}$ on any individual expert probability, and $10^{-8}$ for aggregates.

## 5.4 Parameter learning

For a given iteration of test data, we estimated our model's parameters from sample averages. With $t_0$ instances of $Y = 0$ and $t_1$ instances of $Y = 1$, we use the (Laplace-corrected) estimate $\widehat{p}^o(1) = \frac{t_1 + 1}{t_0 + t_1 + 1}$. By outcome $Y$, we split the continuously-valued covariates into $\mathbf{X}_1$ (a $t_1 \times N$ matrix) and $\mathbf{X}_0$ ($t_0 \times N$). We estimated conditional means by sample averages: $\widehat{\boldsymbol{\mu}}_y^T = \sum_{t=1}^{t_y} \mathbf{X}_y(t, \cdot)$. We then calculated the pooled sample variance $\widehat{\boldsymbol{\Sigma}} = \frac{1}{t_0 + t_1 - 2} \sum_{y=0}^{1} (\mathbf{X}_y - \mathbf{1}\widehat{\boldsymbol{\mu}}_y^T)^T (\mathbf{X}_y - \mathbf{1}\widehat{\boldsymbol{\mu}}_y^T)$.

One difficulty is that estimated weights depend upon the inverse sample covariance matrix, and are sensitive to colinearity. Various strategies to deal with colinearity are available [Panel, 1989], though we do not use them for this demonstration.

For the UCI data sets, we first needed to convert the continuous inputs into log-odds forecasts before applying IndOps or LinOps. As described in section 4.1, our Gaussian model may be used to recalibrate and de-bias individual predictors. Applying this section's estimation procedures to each individual predictor yielded a log-odds forecast for each predictor that was then used in aggregation.

For IndOps we used equation (23) with $\widehat{L}_{1:0}^o = log\frac{\widehat{p}^o(1)}{\widehat{p}^o(0)}$. For LinOps and LogOps, we estimated the parameters in equations (1) and (2) by maximum likelihood (equivalent to choosing $\boldsymbol{w}$ to maximize the log-score on the training data). For AvgOps, we took the equally-weighted arithmetic average of experts' probability forecasts.

### 5.5 Empirical results

Results of experiments on the three simulated Gausian data sets are shown in figure 3, while figure 4 shows results for the UCI data sets. Average log-scores on the test instances are plotted as the number of $N$ randomized experts (predictors) increase.

With simulated Gausian experts, Gaussian and logistic aggregation yielded highest scores in all cases, and were on par with each other. IndOps was equal to Gaussian for $\rho = 0$, but its scores declined precipitously as $\rho$ and $N$ increased. LinOps and AvgOps scored significantly lower than Gaussian aggregation for low $\rho$, but were on par with Gaussian aggregation as $\rho$ increased to 0.50.

## 6    Conclusions

Although the focus of this paper was not on empirical comparisons, we can draw some limited conclusions from the results in section 5.5.

As noted in section 4.3, LinOps aggregates are similar to those of our Gaussian LogOps with high $\rho$. The first experiment highlights that for lower $\rho$, LinOps is unable to fully capture the degree of independence between experts, scoring well below the LogOps (Gaussian and logistic) aggregates.

From just these two UCI datasets in figure 4, it is apparent that the Gaussian model is not dominant in all domains. Yet it does appear to offer better accuracy than IndOps (i.e., when dependence between experts is high). And it can offer much better accuracy than LinOps domains with low expert dependence.

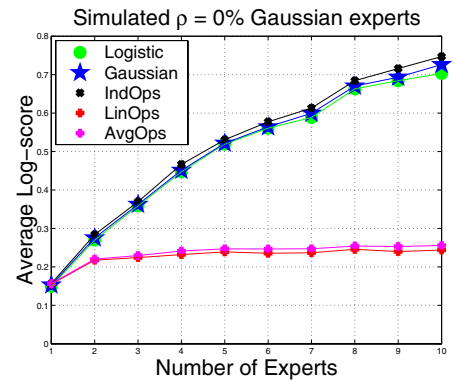

(a) $\rho = 0.0$

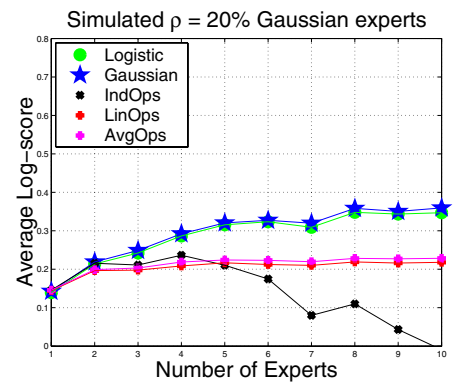

(b) $\rho = 0.2$

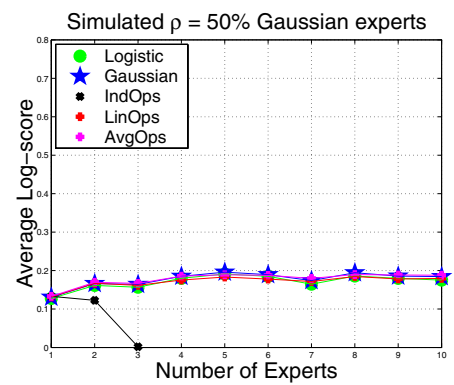

(c) $\rho = 0.5$

Figure 3: Scores for simulated Gaussian experts with Gaussian, Logistic, IndOps, LinOps, and AvgOps aggregation.

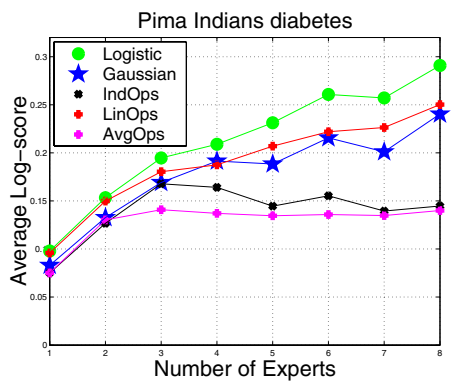

(a) Pima indians diabetes data

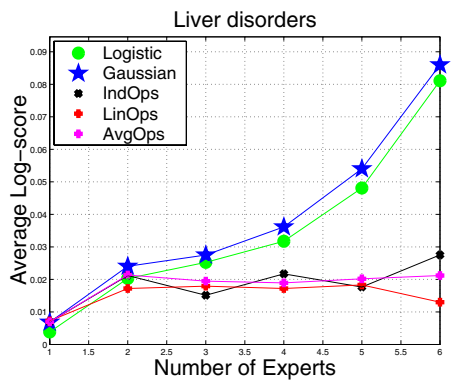

(b) Liver disorder data

Figure 4: Scores for UCI data sets with Gaussian, Logistic, IndOps, LinOps, and AvgOps aggregation.

## 7  Discussion

The generative Gaussian model presented in this paper is essentially one of linear discriminant analysis. This model yields a LogOps result that has the same form as a (discriminative) logistic regression, with empirical results that are similar in many circumstances. The relative merits of generative versus discriminative models — and hybrid approaches — remain an active topic of research [Raina et al., 2004].

Others have observed that overfitting in discriminative models should be tempered by keeping the models simple, and including prior information in estimating aggregation weights on predictors (e.g., within relevance vector machines [Tipping, 2001]). This paper's main contribution is to present a model with analysis that clarifies those factors that directly bear upon the choice of experts' weights: the number of experts, their calibration, dependence, and their relative accu-racy. Using the model's interpretable parameters, a modeler may better understand the impact of these factors when aggregating her information sources.

## Acknowledgements

I wish to acknowledge helpful discussions regarding this research with Ron Howard, Ross Shachter, Peter Morris, and Derek Ayers, as well as constructive comments from anonymous referees.